\documentclass[pdftex,10pt,a4paper]{article} 

\usepackage{amsmath}
\usepackage{amssymb}
\usepackage{times}
\usepackage{graphicx}
\usepackage{color}

\usepackage{hyperref}

\usepackage{setspace}

\title{Echo State Condition at the Critical Point$^1$}

\author{Norbert Michael Mayer\\
Department of Electrical Engineering \\
and \\
Advanced Institute of Manufacturing with High-tech Innovations (AIM-HI), \\
National Chung Cheng University, Chia-Yi, \\
Taiwan\\
mikemayer@ccu.edu.tw
}

\newcommand{\appropto}{\mathrel{\vcenter{
  \offinterlineskip\halign{\hfil$##$\cr
    \propto\cr\noalign{\kern2pt}\sim\cr\noalign{\kern-2pt}}}}}

\begin{document}
\thispagestyle{empty}
\def\YH#1 {{\bf x}_{\it {#1}}}
\def\YI#1 {{\bf u}_{\it {#1}}}
\def\YO#1 {{\bf o}_{\it {#1}}}
\def\YT#1 {{{\bf u}^{\sf teach}_{\it {#1}}}}
\def\YHLIN#1 {{\bf x}_{{\sf lin}, \it {#1}}}

\def \ti {\it t}
\def \tanh {\sf tanh}
\def \sin {\sf sin}
\def \cos {\sf cos}
\def \artanh {\sf artanh}

\def\WWHH {{\bf W}}
\def\WWIH {{\bf w}^{in}}
\def\WWHO {{\bf w}^{out}}

\def\itentitym {{\bf 1}}

\def\gfixp{{{\bf x}_{\infty}}}

\maketitle

\begin{center} { Recurrent networks with transfer functions that fulfill the Lipschitz continuity with $K=1$ may be 
echo state networks if certain limitations on the recurrent connectivity are applied. It has been shown that it is sufficient if the largest singular value of the recurrent connectivity is smaller than 1. 
The main achievement of this paper is a proof under which conditions the network is an echo state network even if the largest singular value is one. It turns out that in this critical case the exact shape of the transfer function plays a decisive role in determining whether the network still fulfills the echo state condition. In addition, several examples with one neuron networks are outlined to illustrate effects of critical connectivity. Moreover, within the manuscript a mathematical definition for a critical echo state network is suggested.}

\end{center}

\footnotetext[1]{This manuscript has been posted at arxiv.org \cite{myselfarxiv}; final paper has been published at\\

{\color{red} \large
\href{http://www.mdpi.com/1099-4300/19/1/3/htm} {Entropy 2017, 19(1), 3; doi:10.3390/e19010003}}

}
\footnotetext[2]{The author declares that he has no conflict of interest.}

\newpage

\section*{Abstract}

\section*{Notations in formulas}

\begin{tabular}[t]{lll}
$s(M)$ &                                          & vector of singular values of matrix $M$ \\ 
$\bf S$  &                                        & diagonal matrix of singular values \\
$s_{\max}$       &                                & largest singular value of a matrix\\
$\lambda(M)$ &                                    & vector of eigenvalues \\
$|\lambda|_{\max}$ &                              & the maximum of the eigenvalues in modulus \\ 
$\theta(.)$    &$\mathbb{R}\rightarrow\mathbb{R}$ & transcendental transfer function \\
$\theta'$, $\dot{\theta}$ &                       & its derivative \\
$\WWIH$        &$\mathbb{R}^{n \times k}$         & input matrix\\
$\WWHH$        &$\mathbb{R}^{k \times k}$         & recurrent transfer matrix \\
$\WWHO$        &$\mathbb{R}^{k \times m}$         & output matrix \\
$\YI { t } $   &$\mathbb{R}^{n}$                  & input time series \\
$\bar{\bf u}^{\infty}$  &                         & complete infinite input time series\\
$\YH { t } $   &$\mathbb{R}^{k}$                  & hidden layer states\\
${\bf y}_{\ti}$&$\mathbb{R}^{k}$                  & alternative hidden layer states to check convergence\\
$\YO { t } $    &                                 & training set \\
$\tilde{\YO { t } }$ &                            & trained network output \\
$d(.,.) $      &$||.||_2$                         & distance measure that is used to check convergence \\
${\bf F}(.)$   &                                  & vector of a set of linearly independent, non-linear functions  \\
${\cal F}_{  \bar{\bf u}^{\infty}}$ & $ \in \cal F$ & dynamics of the network with regard to the input \\
${\cal F}_{crit}$ &                               & all dynamics that show power law forgetting \\
$q_{ \bar{\bf u}^{\infty}, \ti }$&                & time series related to forgetting in ESNs \\
$q_t$ &                                           & time series variable\\
$\phi_1()$  &                                     & cover function to estimate contraction in a 1 neuron net\\
$\phi_k()$  &                                     & dto. for a $k$ neuron network  \\
$\eta$, $\kappa$, $\gamma$     &                  & const parameters of both $\phi_1$ and $\phi_k$ \\
$\Lambda_{\bar{\bf u}^{\infty}}$ &                & Lyapunov exponent of a neural network wrt. an input time series $\bar{\bf u}^{\infty}$
\end{tabular}

\newpage

\section{Introduction}

Classic approaches of recurrent neural networks (RNNs), such as back-propagation through time \cite{werbos1990backpropagation}, have been considered difficult to handle. In particular learning in the recurrent layer is slow  and problematic due to potential instabilities. About 15 years ago reservoir computing \cite{lukovsevivcius2009reservoir} was suggested as an alternative approach for RNNs. Here, it is not necessary to train connectivity in the recurrent layer. Instead, constant, usually random, connectivity weights are used in the recurrent layer. The supervised learning can be done by training the output layer using linear regression. Two types of reservoir computing are well established in the literature. The first is called a liquid state machines (LSMs, \cite{lsms}), which usually bases on a network of spiking neurons. The second type is called an echo state nework (ESN, \cite{jaeger1}), which uses real valued neurons that initially use a sigmoid as a transfer function.
Although a random recurrent connectivity pattern can be used, heuristically it has been found that typically the performance of the network depends strongly on the statistical features of this random connectivity
(cf. for example \cite{theobi2012} for ESNs).

Thus, what is a good reservoir with regard to particular stationary input statistics? This has been a fundamental question for research in this field since the invention of reservoir computing. 
One fundamental idea is that a reservoir can only infer training output from this window of the input history of which traces still can be found inside the reservoir dynamics.
However, if the necessary inference from time series in order to learn the training output is far in the past, it may happen that no traces of this input remain inside the reservoir. 
So, the answer seems to be that a good reservoir is a reservoir from whose states it is possible to reconstruct an input history with a time span that is as long as possible.
More precisely, they should be reconstructed in a way that is sufficiently accurate in order to predict the training output. In other words, a good reservoir is a reservoir that has a good memory of the input history.

There have been efforts to quantify the quality of the memory of the reservoir. Most common is the {\em `memory capacity'} (MC) according to Jaeger's definition \cite{jaeger1}. However, MC has several drawbacks. For example, it is not directly compatible to a Shannon information based measure. Still, it illustrates that ESNs are relatively tightly restricted in the way that the upper limit of the MC is equal to the number of hidden layer neurons. 
So the capabilities of the network increase with the number of neurons.

One more important limiting factor with regard to the reservoir memory is the strength of the recurrent connectivity. According to the echo state condition, the nature of the reservoir requires that the maximum $|\lambda|_{\max}$ of its eigenvalues in modulus is smaller than 1, which is called the echo state property (ESP).
This seems always to result in a exponential forgetting of previous states. 
Thus, forgetting is independent from the input statistics but instead has to be pre-determined and is due to the design of the reservoir dynamics. 


In order to proceed, there are several important aspects. First, it is necessary to get rid of the intrinsic time scale of forgetting that is induced by $|\lambda|_{\max}<1$. More precisely, the remaining activity of inputs to the reservoir that date back earlier than $\Delta t$ is a fraction smaller than $|\lambda|^{\Delta t}_{\max}$. Networks where the largest eigenvalue is larger than $1$ cannot be used as reservoirs anymore, a point which is detailed below.
One can try
$|\lambda|_{\max}=1$ and see if this improves the network performance and how this impacts the memory of the reservoir on earlier events.
Steps toward this direction have been made by going near the "edge of chaos"
\cite{theobi2012} or even further where the network may not be an echo state network for all possible input sequences but instead just around some permissible inputs \cite{manjunath2013echo}. 
Presumably, these approaches still all forget exponentially fast.

Strictly, networks with  $|\lambda|_{\max}=1$ are not covered by the initial proof of Jaeger for the echo state condition. One important purpose of this paper is to close this gap and to complete Jaeger's proof in this sense. 
The other purpose is to motivate the principles of \cite{neco2015} in as simple as possible examples and thus to increase the resulting insight.

The intentions of the following sections of the paper are to motivate the concept of critical neural networks and explain how they are related to memory compression. 
These intentions comprise a  relatively large part of the paper because it seems important to argue for the principle value of critical ESNs.
Sect. \ref{sec:motivation} introduces the concept of reservoir computing and also defines important variables for the following sections. An important feature is that Lyapunov coefficients are reviewed in order to suggest a clear definition for critical reservoirs that can be used analytically on candidate reservoirs. Sect. \ref{sec:create} describes how critical one neuron reservoirs can be designed and also introduces the concept of extending to large networks. Sect. \ref{sec:limit} explains why the critical ESNs are not covered by Jaeger's proof. The actual proof for the echo state condition can be found in sect. \ref{sec:proof}. Certain aspects of the proof have been transferred to the appendix.

\section{Motivation \label{sec:motivation}}
The simplest way to train with data in a supervised learning paradigm is to interpolate data (cf. for example \cite{cs229}). Thus, for a time series of input data $\YI { t } \in \mathbb{R}^n$ that forms an input sequence 
$\bar{\bf u}^{\infty}$  and a corresponding output data 
$\YO { t } \in \mathbb{R}^m$ one can choose a vector of non-linear, linearly independent functions ${\bf F}(\YI{ t } ): \mathbb{R}^n \rightarrow  \mathbb{R}^k$
and a transfer matrix 
$\WWHO: \mathbb{R}^k \rightarrow \mathbb{R}^m $.  Then, one can define

\begin{eqnarray}
\YH { t } &=& {\bf F}(\YI { t } ) \label{eq_interpol}  \nonumber \\
\tilde{\YO { t } } &=& \WWHO \YH { t } . \label{eq_output}  \nonumber
\end{eqnarray}

$\WWHO$ can be calculated by linear regression, i.e.,
\begin{equation}
\WWHO=(AA')^{-1}(AB), \label{linreg}
\end{equation} 
where the rectangular matrices $A=[\YH { 0 }, \YH { 1 }, \dots, \YH { t } ]$ and $B=[{\bf o}_0, {\bf o}_1, \dots, {\bf o}_t]$ are composed from  the data 
of the training set and $A'$ is the transpose of $A$. Further, 
one can use a single transcendental function $\theta(.)$ such that
\begin{equation}
\YH { t } = {\bf F}(\YI { t } ) = \theta(\WWIH \YI { t } ), \label{eq_interpol_2}
\end{equation}
where $\WWIH: \mathbb{R}^n \rightarrow \mathbb{R}^k $ is a matrix in which each line consists of a unique vector and $\theta(.)$ is defined in the Matlab fashion; 
so the function is applied to each entry of the vector separately. Linear independence of the components of $\bf F$ can then be guaranteed if the column vectors of $\WWIH$ are linearly independent. 
Practically, linear independence can be assumed if the entries of $\WWIH$ are chosen randomly from a continuous set and $k\geq n$.

The disadvantage of the pure interpolation method with regard to time series is that the input history, that is $ \YI { t-1 }, \YI { t-2 }, \dots \YI { 0 } $,  has no impact on training the current output 
$\tilde{\YO { t } }$. Thus, if a relation between previous inputs and current outputs exists, that relation cannot be learned.

Different from eq. \ref{eq_interpol}, a {\bf reservoir} in the sense of reservoir computing \cite{jaeger2007special,lukovsevivcius2009reservoir,schrauwen2007overview} can be defined as 
 
\begin{equation}
\YH { t } = {\bf F}(\YH { t-1 }, \YI { t } ) \label{eq_reg}.
\end{equation}

The recursive update function adds several new aspects to the interpolation that is outlined in eq. \ref{eq_interpol}:
\begin{enumerate}
\item The new function turns the interpolation into a dynamical system:
\begin{equation}
\YH { t } = {\bf F}(\YH { t-1 }, \YI { t } ) = {\cal F}_{  \bar{\bf u}^{\infty} }( \YH { t-1 } ),\footnote{For obvious reasons one may call $\bf F$ an input driven system.}
\end{equation}
where the notation ${\cal F}_{  \bar{\bf u}^{\infty} }( \YH { t-1 } )$ is intended to illustrate the character of the particular discrete time, deterministic dynamical system and the fact that each possible time series 
$\bar{\bf u}^{\infty}$ defines a specific set of dynamics. 

The superset ${\cal F}$ over all possible $\bar{\bf u}^{\infty}$, 
\begin{equation}
{\cal F} = \bigcup_{\bar{\bf u}^{\infty}} \; {\cal F}_{  \bar{\bf u}^{\infty} },  \nonumber
\end{equation}
may be called the reservoir dynamics. 
Thus, $\cal F$ covers all possible dynamics of a particular reservoir with regard to any time series of input vectors in $\mathbb{R}^n$.
Note that this way of looking into the dynamics of reservoirs is non-standard. Rather the standard approach is to interpret the reservoir as a non-autonomous 
dynamical system and then to formalize the system accordingly \cite{manjunath2013echo}. For the present work, the turn towards standard dynamical systems has been chosen because here the relevant methodology is well established and the 
above mentioned formalization appears sufficient for all purposes of this work.

\item It is now possible to account for information from a time series' past in order to calculate the appropriate output. 
\end{enumerate}

One important question is if the regression step in the previous section, and thus the interpolation, works at all for the recursive definition in 
eq. \ref{eq_reg}. Jaeger showed  (\cite{jaeger1} pp. 43, \cite{jaeger2}) that the regression is applicable, i.e., the {\bf echo state property} (ESP) is fulfilled, if and only if the network is uniformly state contracting. Uniformly state contraction is defined in the following.

Assume an infinite stimulus sequence $\bar{\bf u}^{\infty} = \left\{ {\bf u}_n \right\}_{n=0}^{\infty}$ 
and two random initial internal states of the system ${\bf x}_0$ and ${\bf y}_0$. 
To both initial states ${\bf x}_0$ and ${\bf y}_0$ the sequences 
$\bar{\bf x}^{\infty} = \left\{ {\bf x}_n \right\}_{n=0}^{\infty}$ and $ \bar{\bf y}^{\infty} =\left\{ {\bf y}_n \right\}_{n=0}^{\infty}$ can be respectively assigned. 
\begin{eqnarray}
{\bf x}_{\ti} = {\bf F}({\bf x}_{\ti-1 },{\bf u}_{\ti } ) &=& {\cal F}_{  \bar{\bf u}^{\infty}} ({\bf x}_{\ti-1 } ) \nonumber \\
{\bf y}_{\ti} = {\bf F}({\bf y}_{\ti-1 },{\bf u}_{\ti } ) &=& {\cal F}_{  \bar{\bf u}^{\infty}} ({\bf y}_{\ti-1 } ) \nonumber \\
q_{ \bar{\bf u}^{\infty}, \ti } &=& d({\bf x}_{\ti }, {\bf y}_{\ti }),
\end{eqnarray}
where $q_{ \bar{\bf u}^{\infty}}$ is another series and $d(.,.)$ shall be a distance measure using the square norm\footnote{Other metric measures may be applicable, though.}.

Then the system ${\cal F}$
is uniformly state contracting if it is independent from $\bar{\bf u}^{\infty}$ and if for any initial state 
(${\bf x}_{0}$,${\bf y}_{0}$) and all real values $\epsilon > 0$ 
there exists  a finite $\tau_{\cal F}(\epsilon)<\infty$ for which 
\begin{equation}
\max_{ \cal F}  q_{ \bar{\bf u}^{\infty}, \ti }  \leq \epsilon \label{metric}
\end{equation}
for all $ {\ti  } \geq {\tau}_{\cal F}$.

Another way to look at the echo state condition is that the network $\cal F$ behaves in a time invariant manner, in the way that some finite subsequence in 
an input time series will roughly result always in the same outcome. In other words
\begin{equation}
{\bf x}_{\Delta t + t_0} \approx {\bf y}_{\Delta t+t_0}  \nonumber
\end{equation}
independent of $t_0$, ${\bf x}_{t_0}$ and ${\bf y}_{t_0}$ and if $\Delta t$ is sufficiently large.

{\bf Lyapunov analysis} is a method to analyze predictability versus instability of a dynamical system (see \cite{wainrib2016local}). 
More precisely, it measures exponential stability.

In the context of non-autonomous systems, one may define the Lyapunov exponent as
\begin{equation}
\Lambda_{\bar{\bf u}^{\infty}} = \lim_{|q_{ \bar{\bf u}^{\infty}, \ti=0 }|\rightarrow 0 } \; \; \lim_{t \rightarrow \infty} \frac{1}{t} \log \frac{|q_{ \bar{\bf u}^{\infty}, \ti }|}{|q_{ \bar{\bf u}^{\infty}, 0 }|},  
\label{lyapunov_exp_eq}
\end{equation}
Thus, if 
\[ q_{ \bar{\bf u}^{\infty}, \ti} \appropto \exp(b \ti), \]
then $\Lambda_{\bar{\bf u}^{\infty}}$ approximates $ b $ and thus measures the exponent of exponential decay. For power law decays, the Lyapunov exponent is always zero.
\footnote{For example one may try $q_{ \bar{\bf u}^{\infty}, \ti} = \frac{1}{\ti +1}$. }

In order to define criticality, we use the following definition.

{\bf Definition.} A reservoir that is uniformly state contracting shall be called {\bf critical} if for at least one input sequence ${\bar{\bf u}^{\infty}}$ there is at least one Lyapunov exponent
$\Lambda_{\bar{\bf u}^{\infty}}$ that is zero.

The {\bf echo state network} (ESN) is an implementation of reservoir dynamics as outlined in eq. \ref{eq_reg}. 
Like other reservoir computing approaches, the system is intended to resemble the dynamics of a biologically 
inspired recurrent neural network.
The dynamics can be described for discrete time-steps $\ti$, with the following equations:
\begin{eqnarray}
\YH { lin, \ti} &=& { \WWHH \YH { \ti-1 } } + \WWIH \YI { \ti } \label{xlin}\\
\YH { \ti} &=& \theta \left(   \YH { lin, \ti}     \right) \nonumber \\
\tilde{\YO { t } } &=& \WWHO \YH {\ti }. \nonumber
\label{hidden_dyn} 
\end{eqnarray}

With regard to the transfer function $\theta(.)$, it shall be assumed that it is continuous, differentiable, transcendental and monotonically increasing with the limit
$1 \geq \theta'(.) \geq 0$, which is compatible with the requirement that $\theta(.)$ fulfills the Lipschitz continuity with $K=1$.
Jaeger's approach uses random matrices for $\WWHH$ and $\WWIH$, learning is restricted to the output layer $\WWHO$. 
The learning (i.e., training $ \YO { \ti } $) can be performed by linear regression (cf. eq. \ref{linreg}). 

The ESN fulfills the echo state condition (i.e., it is uniformly state contracting) if certain restrictions on the connectivity of the recurrent layer apply, for which one can name a necessary condition and a sufficient condition:
\begin{itemize}
\item {\bf \sf C1} A network has echo states only if 
\begin{equation}
1 > |\lambda|_{\max} = \max {\rm abs}(\lambda (\WWHH)), \label{necessary_eq}
\end{equation}
i.e. the absolute value of the biggest eigenvalue of $\WWHH$ is below $1$. The condition means that a network is {\bf not} and ESN if $1 < \max {\rm abs}(\lambda (\WWHH))$.
\end{itemize}

\begin{itemize} 
\item {\bf \sf C2} 
Jaeger named here initially
\begin{equation}
1 > s_{\max} = \max s (\WWHH ), \label{sufficient_eq_old}
\end{equation}
where $s$ is the vector of singular values of the matrix $\WWHH$.
However, a closer sufficient condition has been found in \cite{tighter}. Thus, it is already sufficient to find a full rank matrix $D$ for which
\begin{equation}
1 > \max s (D \WWHH D^{-1}). \label{sufficient_eq}
\end{equation}
\end{itemize}
\cite{tighter2} found another formulation of the same constraint:
{ \em The network with internal weight matrix $\WWHH$ satisfies the
echo state property for any input if W is diagonally Schur stable, i.e., there exists a diagonal
$P > 0$ such that $W^T P W-P$  is negative definite.}

Apart from the requirement that a reservoir has to be uniformly state contracting,
the learning process itself is not of interest in the scope of this paper.

\section{Critical reservoirs with regard to the input statistics $\bar{\bf u}^{\infty}$ \label{sec:create}}
Various ideas on what types of reservoirs work better than others have been brought up. 
One can try to keep the memories of the input history in the network as long as possible. The principle idea is to tune
the network's recurrent connectivity to a level where 
the convergence for a subset of ${\cal F}_{crit} \in {\cal F} $ with regard to eq. \ref{metric} is
\begin{equation}
q_{ \bar{\bf u}^{\infty}_{crit}, \ti } \appropto t^{a} \label{powerlaw}
\end{equation}
rather than 
\begin{equation}
q_{ \bar{\bf u}^{\infty}, \ti } \appropto b^t, \label{exponential}
\end{equation}
where $a<0$ and $0<b<1$ are system specific values, i.e., they depend on ${\cal F}_{  \bar{\bf u}^{\infty}}$. 

A network according to eq. \ref{powerlaw} is still an ESN since it fullfils the ESP. Still, forgetting of initial states is not bound to a certain time scale. Remnants of information can 
--under certain circumstances-- remain for virtually infinite long times within the network given that not too much unpredictable new input enters the network. 
Lyapunov analysis of a time series according to eq. \ref{powerlaw} would result in zero, and Lyapunov analysis of eq. \ref{exponential} yields a nonzero result.

In ESNs forgetting according to the power law of eq. \ref{powerlaw} for an input time series $q_{ \bar{\bf u}^{\infty}_{crit}, \ti }$ is achievable if the following constraints are fulfilled
:
\begin{itemize}
\item The recurrent connectivity $\WWHH$, the input connectivity $\WWIH$ of the ESN and the transfer function $\theta(.)$ have to be arranged in a way that if the ESN is fed with $q_{ \bar{\bf u}^{\infty}_{crit}, \ti }$
one approximates
\begin{equation}
\lim_{\ti \rightarrow \infty} \left|\dot{\theta}(\YH { lin, \ti }) \right|=1. \nonumber
\end{equation}
Thus, the aim of the training is
\begin{equation}
\left|\dot{\theta}(\YH { lin, \ti }) \right|=1. \label{crit_update}
\end{equation}
Since the ESN has to fulfill the Lipschitz continuity with $K=1$, the points where $\dot{\theta}=1$ have to be inflection points of the transfer function. In the following these inflection points 
shall be called epi-critical points (ECPs).

\item The recurrent connectivity of the network is to be designed in a way that the largest absolute eigenvalue and the largest singular value of $\WWHH$ both are equal to one. This can be done by using 
normal matrices for $\WWHH$ (see sect. \ref{sec:normal}).

\end{itemize}

\begin{figure}[t]
\begin{center}
\includegraphics[  width=0.28\paperwidth]{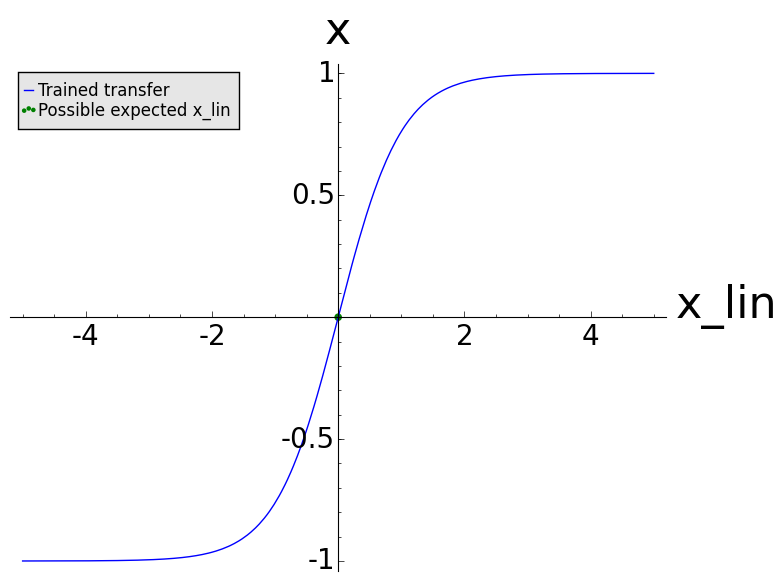}
\includegraphics[  width=0.28\paperwidth]{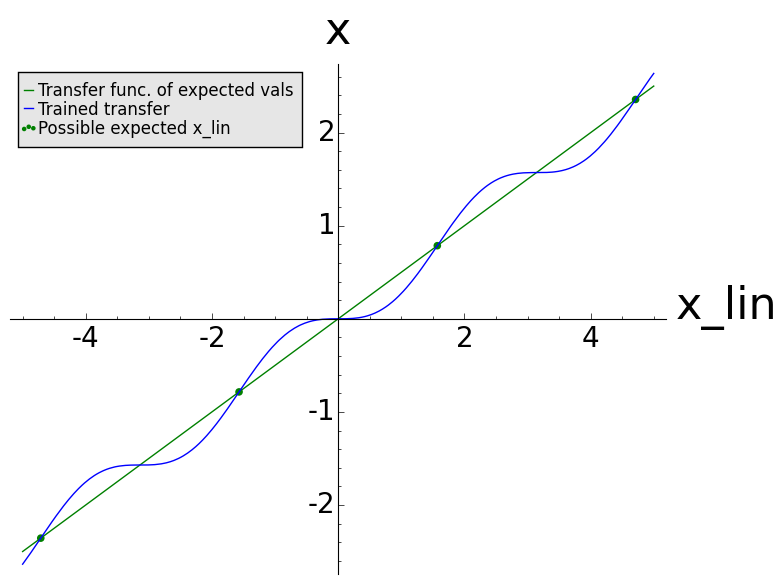}
\end{center}
\caption{\label{transfer_functions} Variant possible transfer functions with interesting features regarding criticality. On the left side $\tanh$ has only one epi-critical point. Eq. \ref{sigmoid} is graphed on the right side. Here, there is an infinite set of epi-critical points $n \pi + \pi/2$ that are all positioned along the line $y=x/2$. In both graphs green dots indicate epi-critical points, green curves are smooth interpolations between those points, and the blue line indicates 
the particular transfer function itself. }
\end{figure}

\begin{figure}[t]
\begin{center}
\includegraphics[  width=0.38\paperwidth]{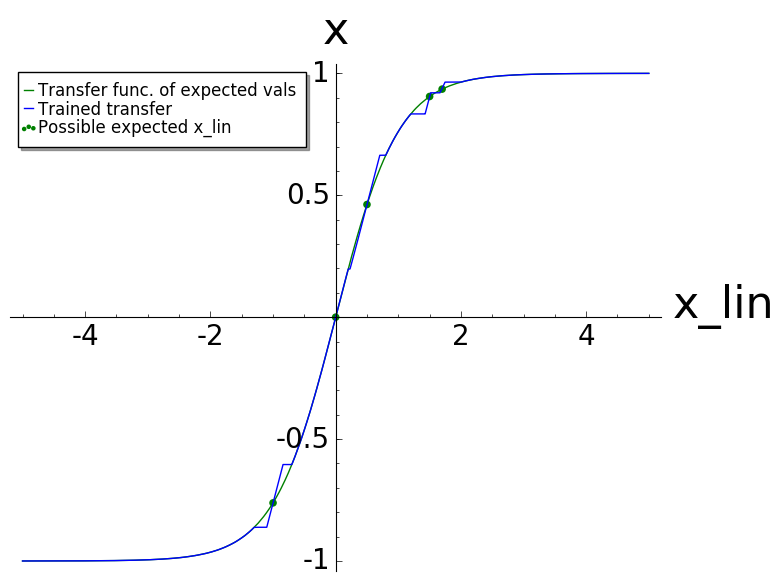}
\end{center}
\caption{\label{tailored_transfer} 
Tailored transfer function where the epi-critical points can be organized in an adaptive way and follow a transcendental function (which has certain advantages to the examples depicted in fig. \ref{transfer_functions}, in blue).
Green dots indicate epi-critical points, the green curve is a smooth interpolation between those points, and the blue line indicates the particular transfer function itself. Note that the green curve is a $\tanh$ and thus a transcendental function.
}
\end{figure}

\subsection{Transfer functions\label{anticip_sect}}

The standard transfer function that is widely used in ESNs and other types of neural networks is the sigmoid function, i.e.,
\begin{equation}
\theta(x)={\tanh}(x). \label{tanh}
\end{equation} 
The sigmoid transfer function has one ECP $\Pi_0=0$.

As a second possible transfer function, one may consider
\begin{equation}
\theta(x) = 0.5 x - 0.25 \, {\sin}(2 x).  
\label{sigmoid}
\end{equation}
Here, one has a infinite set of ECPs at $\Pi_i=(n+0.5)\times \pi$.
It is important to have more than one ECP in the transfer functions because
for a network with discrete values it appears necessary that each neuron has at least 2 possible states in order to have any inference from the past to the future. 
In the case of $\tanh$, the only solution of eq. \ref{crit_update} is
\[
\YH { lin, \ti } = 0.
\]

That type of  trained network cannot infer information from the past to the present for the expected input, which significantly restricts its capabilities. One can see that from
\begin{equation}
\YH { t } = \tanh ( \YH { lin,t }  ) = 0.  \nonumber
\end{equation}
The next iteration yields
\begin{equation}
\YH { t+1 } = \tanh (\WWIH \YI { \ti } ),  \nonumber
\end{equation}

which again, after training would require
\begin{equation}
0=\WWIH \YI { \ti } .  \nonumber
\end{equation}
Thus, the network can only anticipate the same input at all time steps.

In the case of eq. \ref{sigmoid}, 
the maximal derivative $\theta'(x)$ is $1$ at $x = \pi(n+1/2)$, where $n$ is an integer number (confer fig. \ref{transfer_functions}).
Here the main advantage is that there exists an infinite set of epi-critical points. However, all these points are positioned along 
the linear function $y=x/2$. This setting still significantly restricts the training of ${\cal F}_{crit}$.
Here one can consider the polynomial with the lowest possible rank (cf. the green line in 
fig. \ref{transfer_functions}, left side) 
that interpolates between the epi-critical points (in the following called an epi-critical transfer function). In the case of eq. \ref{sigmoid} 
the epi-critical transfer function is the linear function
\begin{equation}
\YH { t+1 } = 0.5 \WWHH \YH { \ti } + 0.5 \WWIH \YI { \ti }.  \nonumber
\end{equation}
Thus, the effective dynamics of the trained reservoir on the expected input time series is -if this is possible- the dynamics of a linear network. This results in a very restricted set of 
trainable time series.

As an alternative, one could consider also a transcendental function for the interpolation between the points, such as depicted in fig. \ref{tailored_transfer}. The true transfer function (blue line in fig. 
\ref{tailored_transfer}) can be constructed in the following way.
Around a set of defined epi-critical points $\Pi_i$, define $\theta$ as either
\begin{equation}
\theta(x) = {\tanh}(x-\Pi_i)+{\tanh}(\Pi_i)  \nonumber
\end{equation}
or
\begin{equation}
\theta(x) = {\tanh}(x).  \nonumber
\end{equation}
This is one conceptional suggestion for further investigations. The result is a transfer function with the epi-critical points $\Pi_i$ and $0$. The epi-critical transfer in this case is a $\tanh$ function.

\subsection{Examples using a single neuron as a reservoir}
In this and the following sections, practical examples are brought up where a single neuron represents a reservoir. Single neuron reservoirs have been studied in other researches \cite{appeltant2011information,ortin2015unified}. 
Here the intention is to illustrate the principle benefits and other features of critical ESNs.

First one can consider a neuron with $\tanh$ as a transfer function along with a single input unit 
\begin{equation}
x_{t+1} = {\tanh}(b x_t + u_t), \label{single1}
\end{equation}
where in order to achieve a critical network $|b|$ has to be equal to 1.
I.e., the network exactly fulfills the boundary condition. 
From previous consideration 
one knows that ${\cal F}_{crit}$ has the dynamics that results from $u_t=0$ as an input.  In this case 
the only fixed-point of the dynamics is $x_t=0$, which is also the epi-critical point if $|b|=1$.

{\bf Power law forgetting}: Starting from the two initial values $x_0=0$ and say $y_0=0.01$, one can see that the two networks converge in a power law manner to zero. 
On the other hand, for a linear network
\begin{equation}
x_{t+1} = b x_t + u_t  \nonumber
\end{equation} 
with the same initial conditions the dynamics of the two networks never converge (independently of $u_t$). Instead, the difference between $x_0=0$ and say $y_0=0.01$ stays the same forever. 
Thus, the network behavior in the the case of $|b|=1$ depends on the nature of the transfer function.
For all other values of $b$, both transfer functions result qualitatively in the same behavior in dependence on $b$: either they diverge or they converge exponentially. 
Since $|b|=1$ is also the border between convergence and divergence and thus the border between uniformly state contracting networks and not uniformly state contracting networks, the case of $|b|=1$ is a critical point of the dynamical system, in a similar manner as a critical point at the transition from ordered dynamics to instability. 
In the following it is intended to extend rules for different transfer functions, where different transfer functions result in the critical point in 
uniformly state contracting networks and where this is not the case. 

As a final preliminary remark, it has to be emphasized that a network being uniformly state contracting means that the states are contracting for 
any kind of input $u_t$. It does not mean that for any kind of input the contraction follows a power law. In fact, for all input settings  
$u_t \neq 0$ the contraction is exponential for the neuron of eq. \ref{single1},
even in the critical case ($|b|=1$).

\begin{figure}[t]
\begin{center}
\includegraphics[  width=0.38\paperwidth]{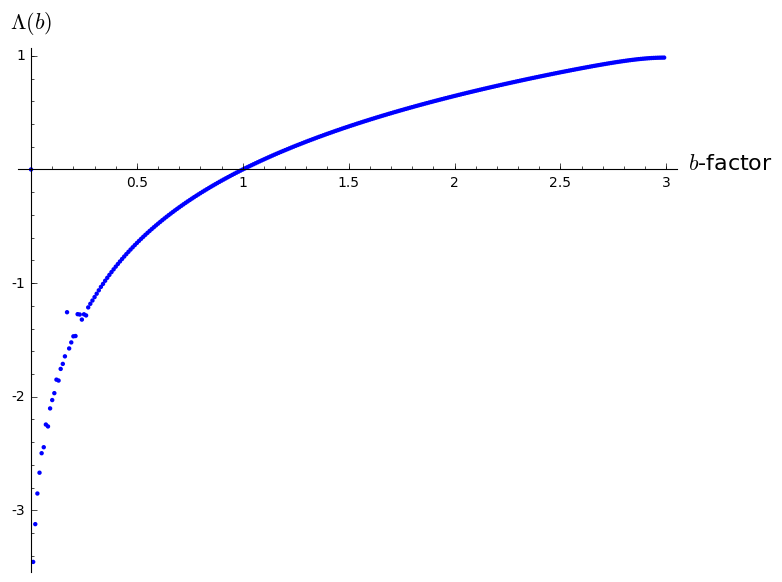}
\end{center}
\caption{\label{fig:lyapunov} Depicted is the Lyapunov exponent for the example system of eq. \ref{alternating} for different values of $b$. At $b=1$ the Lyapunov exponent crosses zero. 
At this point all types of linear analysis fail. The point marks the border between networks that fulfill the ESP and those that do not. 
So, this point is the called critical point.  Further analysis shows that the point itself belongs to the area with ESP. All results from figures \ref{one_neuron_fig} and \ref{one_neuron_fig2} are drawn from the point where the Lyapunov exponent is zero, that is $b=1$.}
\end{figure}

\begin{figure}[t]
\centerline{
\includegraphics[  width=0.5\paperwidth]{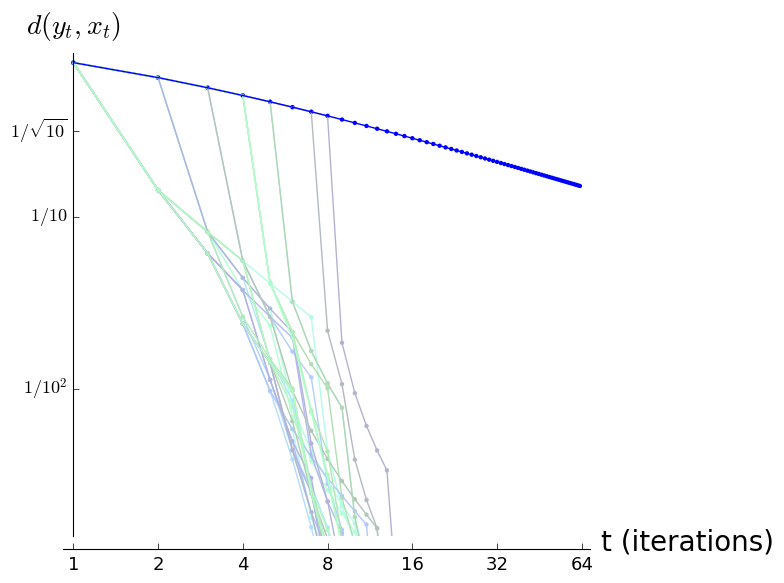}}
\caption{\label{one_neuron_fig} Results from two dynamics defined in eq. \ref{alternating}, with $b=1$ and different types on input ${\bf u}_t$. 
One copy of the two initially identical networks has received a variant input at iteration one.
Depicted here is the  decay of the difference of the state in the recurrent layer if both networks receive the same and expected (alternating) input (in dark blue). 
The pale curves are data from fig. \ref{one_neuron_fig2} embedded for comparison. One can see that the difference function (blue) follows a power law and pertains for longer than 64 iterations.
}
\end{figure}

\begin{figure}[t]
\centerline{
\includegraphics[  width=0.3\paperwidth]{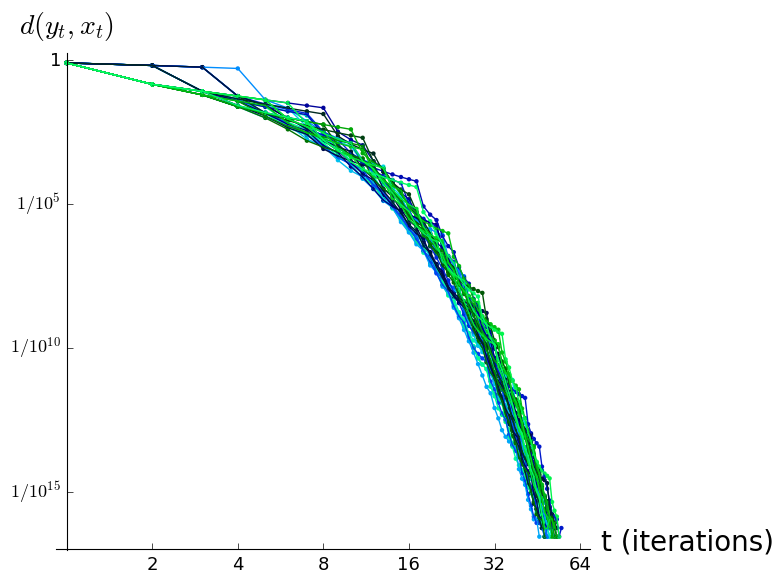}
\includegraphics[  width=0.3\paperwidth]{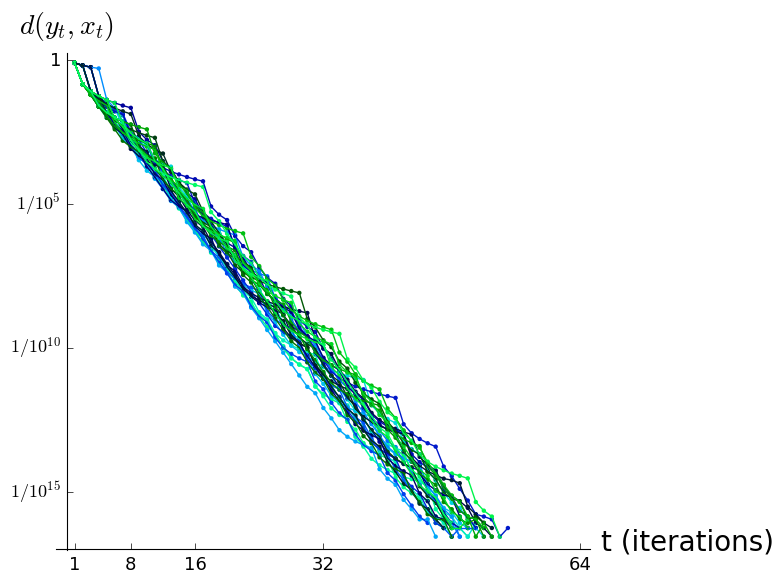}}
\caption{\label{one_neuron_fig2} Complementary data for fig. \ref{one_neuron_fig}:
Left: Decay if both networks receive the same but irregular (i.i.d random) input (different colors are used 
for different trials). The difference vanishes much faster than in the left case. Finally the difference becomes and stays null, which is out of a logarithmic scale. Thus, later iterations are not depicted anymore. Right: Depicted here is the same data as at the left, however in a double-logarithmic plot. The data forms a straight line, which indicates an exponential decay.}
\end{figure}

\subsection{Single neuron network example with alternating input and power law forgetting}
As outlined above,
for practical purposes the sigmoid function, i.e., $\theta=tanh()$, is not useful for critical networks because 
the only critical state occurs when total activity of such a network is null. In that case it is not possible 
to transfer information about the input history. 
The reason is illustrated in the following example, where instead of a sigmoid function other types of  transfer functions 
are used.

So, the one neuron network  
\begin{equation}
x_{t+1} = \theta(-b x_{t} + (2-b) u_t) \label{alternating}
\end{equation}
with a constant $b$, the transfer function of eq. \ref{sigmoid} and the expected alternating input $u_t=(-1)^t \pi/4$ has an attractor state when also $x_t$ is alternating with $x_t=(-1)^t \pi/4$ independently from $b$. Thus, $u_t=(-1)^t \pi/4$ shall be interpreted as the expected input that directs the activity of the network exactly to where $\theta'=1$ and thus induces a critical dynamic ${\cal F}_{\bar{\bf u}^{\infty}} = {\cal F}_{crit}$. 

It is now interesting to investigate the convergence behavior for different values of $b$ considering differing starting values for internal states $x_t$ and $y_t$. Fig. \ref{fig:lyapunov} depicts the resulting different Lyapunov  exponents for one neuron networks with different values of $b$. One can see that --not surprisingly-- the Lyapunov exponent for $b=1$ is zero. This is the critical point that marks the transition from order to instability in the system and at the same time the transition from ESP to non-ESP.

In the following the same network at $b=1$ is investigated. Some unexpected input, i.e., $u_t\neq(-1)^t \pi/4 $, lets the network jump out of the attractor. If the input afterwards continues as expected, the network slowly returns to the attractor state in a power law fashion. Thus,
\begin{equation}
d(x_t, y_t) \appropto t^{-a},  \nonumber
\end{equation}
where $y_t$ represents the undisturbed time series and $a$ is a constant value. Note that $x_t$ contains all information of the network history and that the network was simulated using 
IEEE 754-2008 double precision variables, which have a memory size of 64 bits on Intel architecture computers. Although floating point variables are organized in a complicated way of three parts, the sign, exponent and mantissa, it is clear that the total reservoir capacity cannot exceed those 64 bits, which means in the limit a reservoir of one neuron cannot remember more than 64 binary i.i.d. random numbers.

Thus, about 64 iterations after an unexpected input the difference 
between $x_t$ and $y_t$ should be annihilated. Thus, if both networks $x_t$ and $y_t$ receive the same unexpected input that is of the same magnitude, the difference
between the $x_t$ and $y_t$ should reach virtually $0$ within 64 iterations. The consideration can be tested by setting the input
\begin{equation}
u_t = \pi/4 \times {\rm rand},  \nonumber
\end{equation}
where $\rm rand$ is an i.i.d. random list of +1s and -1s,
that produces a representative of ${\cal F}_{\bar{\bf u}^{\infty}} \neq {\cal F}_{crit}$. Fig. \ref{one_neuron_fig} and fig. \ref{one_neuron_fig2} depict
results from simulations, where after one iteration when two networks receive different inputs, both networks receive again the same input. Depicted is again the development of the difference between both networks $d(x_t,y_t)$ versus the number of iterations. The graphs appear in a double logarithmic fashion, so power law decays appear as straight lines. One can see that the networks that receive alternating, i.e. the expected input, pertains the difference for very long time spans (that exceed 
64 iterations). On the other hand, if the network input for both networks is identical but i.i.d random and of the same order of magnitude as the expected input, the difference vanishes within 64 iterations. 

The network distinguishes regular input from irregular input. Memories of irregular events pertain for a long time in the network provided that the following input is regular again. How a reservoir can be trained to anticipate certain input statistics has been discussed \cite{neco2015}. Additional solutions to this problem are subject to further investigations.
 
\subsection{Relation to {\em `near edge of chaos'}  approaches}
It is a common experience that -- in spite of given theoretical limits for the ESPs -- the recurrent connectivity can be significantly higher than $s_{\max}=1$ for many practical input statistics. 
Those over-tuned ESNs in many cases show a much better performance than those that actually obey Jaeger's initial limit ESP. 
So, recently researchers came up with theoretical insights with regard to ESPs that are subject to a network and a particular input statistic~\cite{manjunath2013echo}. 
In the scope of this work, instead of defining the ESP for a network the ESP is always defined as related to a network and an input statistic.
Also, similar efforts have been undertaken in the field of so-called liquid state machines \cite{2005natschlaeger,legenstein2007edge}.

One may assume that those approaches show similar properties as the one that has been presented here. 
However, for a good reason those approaches all are called {\em `near edge of chaos'} approaches.
In order to illustrate the problems that arise from those approaches, one may consider what happens if those overtuned ESNs are set exactly to the critical point. Here, just for the general understanding one may
consider again a one neuron network and a tanh as a transfer function, so
\begin{equation}
x_{t+1} = {\tanh}(-b x_{t} + u_t). \label{alternating_old}
\end{equation}
Note that the ESP limit outlined above requires that the recurrent connectivity should be $b<1$. 
An input time series than one can use is from the previous section
\begin{equation}
u_t=(-1)^t \pi/4 \label{input_ts}
\end{equation}
Slightly tedious but basically simple calculus results in a critical value of $b\approx2.344$ for the input time series, where 
$x_t \approx (-1)^t \times 0.757$. In this situation one can test for the convergence of two slightly different initial conditions and obtain a power law decay of the difference. 
However, setting up the amplitude of the input just a tiny bit higher is going to result in two diverging time series $x_t$ and $y_t$. If the conditions of the ESN are chosen to be 
exactly at the critical point, it is possible that a untrained input sequence very close to the trained input sequence turns the ESN into a state where the ESP is not fulfilled anymore
(for a related and more detailed discussion with a numerical analysis, confer \cite{2016antici}).

For this reason all the networks have to be chosen at a significant margin away from the edge of instability. That is very different from the 
approach in the previous section where although the expected input sequence for the network is exactly at the critical point, all other input sequences result in a 
stable ESN in most cases with exponential forgetting.

\subsection{How the one neuron example can be extended to multi-neuron networks: Normal Matrices \label{sec:normal}}

A normal matrix $\WWHH$  commutes with its own transpose, i.e.,
\begin{equation}
\WWHH \WWHH^T = \WWHH^T \WWHH.  \nonumber
\end{equation}
For a normal matrix $\WWHH$, it can easily be shown that
\begin{equation}
 \max s (\WWHH) = \max {\rm abs}(\lambda (\WWHH)).  \nonumber
\end{equation}
These matrices apply to the spectral theorem; the largest absolute eigenvalue is the same as the largest singular value, which makes a clear theoretical separation between 
networks that are uniformly state contracting and those that are not compatible to the echo state condition. Still, for normal matrices all previously known ES conditions do not determine to which of those two groups the critical point\footnote{Which is a set of Lebesgue measure null.} itself belongs.

Summarizing, all previous works result in theorems for an {\bf open set} of conditions that are defined by the strict inequalities eqs. (\ref{necessary_eq}) and (\ref{sufficient_eq}). In the 
closest case, the case of {\bf  normal} matrices, when considering the singular condition
\begin{equation}
1 {\bf =} \max s(\WWHH) = \max {\rm abs}(\lambda (\WWHH)) \label{boundary}
\end{equation}
there is no statement of the above mentioned theorems if the network is uniformly state contracting. 

Some simple, preliminary numerical tests reveal that in the case 
of networks that satisfy eq. \ref{boundary} the further development of the network strictly depends on the exact shape of transfer function.

\section{Echo state condition limit with weak contraction\label{sec:limit}}

In the previous section, it has been shown how power law forgetting may occur in an ESN type neural network. These networks are all tuned to the point where $|\lambda|_{\max}=S=1$. For this tuning it is still undetermined if the ESP is fulfilled or not even if normal matrices or one neuron RNNs are used. The current section is dedicated to determining under which conditions Jaeger's ESP can be extended to this boundary condition.

Jaeger's sufficient echo state condition (see \cite{jaeger1}, App. C, p. 41) has strictly been proven only for non-critical systems (largest singular value $S<1$)  and with $\tanh(.)$ as a transfer function. The original proof is based on the fact that $\tanh$ in combination with $S<1$ is a contraction. In that case Jaeger shows an exponential convergence.

The core of all considerations of a sufficient condition is to give a close upper estimate of the distance between the next iterations of two different states 
$y_t$ and $x_t$. The estimate is of the form 
\begin{equation}
\max_{\bar{\bf u}^{\infty}}  d(y_{t+1}, x_{t+1}) = \max_{\cal F} d({\cal F}(y_t),{\cal F}(x_t)) \leq \phi_1 \, \cdot \, d(y_{t}, x_{t}),  \nonumber
\end{equation}
where the parameter $\phi_1$ basically is quantified by the nature of the transfer function and the strength of the connectivity matrix. The estimate has to be good enough that the  iterative application 
of $\phi_1$ should result in a convergence to $0$:
\begin{equation}
\lim_{t\rightarrow\infty} [\, \phi_1 \, \cdot \, ]^t d (y_0,x_0) =0, \label{lim_conv}
\end{equation}
This is equivalent to investigating a series $q_t$ with
\begin{equation}
d(y_t, x_t)  < q_t =  [\, \phi_1 \, \cdot \, ]^t d (y_0,x_0)  \nonumber
\end{equation}
and 
\begin{equation}
\lim_{t\rightarrow\infty} q_t =0, \label{lim_conv2}
\end{equation}

which can prove that the requirement of uniformly state contraction (cf. eq. \ref{metric}) is fulfilled. 
For example, consider the case of a reservoir with one neuron as described in eq. \ref{single1}.
Here the challenge is to find an estimator for $\phi_1$ such that
\begin{equation}
\max_{u_t \in \mathbb{R}} d(\theta( b y_t + u_t), \theta( b x_t + u_t )  )  \leq \phi_1 \, \cdot \, d(y_{t}, x_{t}),  \nonumber
\end{equation}
where the chosen $\phi_1$ still holds the limit in eq. \ref{lim_conv}.
For $|b|<1$, convergence can be proven easily:
\begin{eqnarray}
\max_{u_t \in \mathbb{R}} d(\theta( b y_t + u_t), \theta( b x_t + u_t )  )  &\leq&  \nonumber\\
\max_{u_t \in \mathbb{R}} || b y_t + u_t - b x_t - u_t ||_2                 & =&  \nonumber\\
|b| \, \cdot \,  d(y_t, x_t).  \nonumber
\end{eqnarray}
Thus for $|b|<1$, one can easily define $\phi_1=|b|$. So $q_t$ can be defined as
\begin{equation}
q_t = |b|^t \, \cdot \,  d(y_0, x_0).   \nonumber
\end{equation}
So eq. \ref{lim_conv2} is fulfilled. The convergence is exponential. 
The arguments so far are analogous to the core of Jaeger's proof for the sufficient echo state condition
{\sf C2} that is restricted to one dimension. 

For $|b|=1$, this argument does not work anymore. 
Obviously, Jaeger's proof is not valid under these circumstances.
However, the initial theorem can be extended. As a pre-requisite, one can replace the constant $\phi_1$ with a function 
that depends on $q_t$ as an argument. 

So one can try
\begin{equation}
\phi_1(q_t) = 1 - \eta q_t^\kappa,  \nonumber
\end{equation}
where $\eta>0$ and $\kappa>1$ have to be defined appropriately.
This works for small values of $q_t$. However, it is necessary to name a limit for large $q_t>\gamma$. Define
\begin{equation}
\phi_1(z) := \begin{cases}  1-\eta z^{\kappa} & \mbox{if } z < \gamma  \\ 1-\eta \gamma^{\kappa} & \mbox{if } z \geq \gamma. \end{cases}
\label{weakcontrf1}
\end{equation}

Three things have to be done to check this cover function:
\begin{itemize}
\item First of all, one needs to find out if indeed the cover function $\phi_1$ fulfills
\begin{equation}
\max_{u_t \in \mathbb{R}} d(\theta(  y_t + u_t), \theta(  x_t + u_t )  )  \leq d(y_t, x_t) \phi_1(d(y_t, x_t)).  \nonumber
\end{equation}

In order to keep the proof compatible with the proof for multiple neurons for this work, one has to chose a slightly different
application for $\phi_1$,
\begin{equation}
\max_{u_t \in \mathbb{R}} d^2(\theta(  y_t + u_t), \theta(  x_t + u_t )  )  \leq d^2(y_t, x_t) \phi_1 (d^2(y_t, x_t)),  \nonumber
\end{equation}
which serves the same purpose and is much more convenient for multiple neurons.

In app. \ref{app_contraction_func} one can find a recipe for this check.

\item Second, one has to look for the convergence of   
\begin{equation}
q_{t+1} = q_t \, \phi_1(q_t) = q_t \, (1-  \eta q_t^\kappa),  \nonumber
\end{equation}
when $q_t \leq \gamma$. 
The analysis is done in app. \ref{app_sequence_1}.
\item Third, one needs to check
\begin{equation}
q_{t+1} = q_t \, \phi_1(q_t) = q_t \, (1-  \eta \gamma^\kappa),  \nonumber
\end{equation}
as long as $q_t>\gamma$. Since the factor
$(1-  \eta \gamma^\kappa)$ is positive, smaller than one and constant, the convergence process is exponential, obviously.
\end{itemize}
Note that the next section's usage of the cover function differs slightly even through it has the same form as eq. \ref{weakcontrf1}.

\section{Sufficient condition for a critical ESN\label{sec:proof}}
The content of this section is a replacement of the condition {\sf C2} where the validity of the ESP is inferred for 
$S\leq1$.

{\bf Theorem:} If hyperbolic tangent or the function of eq. \ref{sigmoid} are used as transfer functions,
the echo state condition (see eq. \ref{metric}) is fulfilled even if $S=1$.

{\bf Summary of the proof:}
As an important precondition, the proof requires that both transfer functions fulfill
\begin{equation}
d(\theta(y_t),\theta(x_t)) \leq d(y_t, x_t) \phi_k (d^2(y_t, x_t)), \label{defweak}
\end{equation}
where $\phi_k(z)$ is defined for a network with $k$ hidden neurons as
\begin{equation}
\phi_k(z) := \begin{cases}  1-\eta z^{\kappa} & \mbox{if } z < \gamma  \\ 1-\eta \gamma^{\kappa} & \mbox{if } z \geq \gamma \end{cases}.
\label{weakcontrf}
\end{equation}
Here, $1 > \gamma > 0, 1 > \eta > 0$, $\kappa \geq 1$ are constant parameters that are determined by the transfer function and the metric norm $d(.,.)=||.||_2$

In app. \ref{app_contraction_func} it is shown that indeed both transfer functions fulfill that requirement.
It then remains to prove that in the slowest case we have a convergence in a process with 2 stages. In the first stage, if $d^2(y_t,x_t) > \gamma $ there is a convergence that 
is faster or equal to an exponential decay. The second stage is a convergence process that is faster or equal to a power law decay.

{\bf Proof:} Note with regard to the test function $\phi_k$:
\begin{eqnarray}
&& \phi_k \leq 1,\nonumber \\
\forall z, \forall Z: 0 \leq z \leq Z & \leftrightarrow & \phi_k(z) \geq \phi_k (Z), \nonumber \\
\mbox{and} \; \; \forall z, \forall Z: 0 \leq z \leq Z & \leftrightarrow & Z \times \phi_k(Z) \geq z \times \phi_k (z) \nonumber 
\end{eqnarray}  

In analogy to Jaeger, one can check now the contraction between the time step $t$ and $t+1$:
\begin{eqnarray} 
d^2({\bf y}_{t+1},{\bf x}_{t+1}) &=& d^2(\theta({\bf y}_{lin,t+1}),\theta({\bf x}_{lin,t+1})) \nonumber \\
&\leq & d^2({\bf y}_{lin,t+1}, {\bf x}_{lin,t+1}) \nonumber \\
&&\times \phi_k (d^2({\bf y}_{lin,t+1},{\bf x}_{lin,t+1}))
\label{jaeger1}
\end{eqnarray}
One can rewrite 
\begin{eqnarray}
d^2({\bf y}_{lin,t+1}, {\bf x}_{lin,t+1}) &=& \nonumber \\
||  \WWHH {\bf y}_{t} + I -  \WWHH {\bf x}_{t} - I ||^2_2 &=& \nonumber \\
|| \WWHH ( {\bf y}_{t} - {\bf x}_{t} ) ||^2_2,
\end{eqnarray}
where $I= \WWIH {\bf u}_{t}$. Next one can consider that one can decompose the recurrent matrix by using singular value decomposition (SVD) and obtain
$\WWHH = {\bf U} \cdot {\bf S} \cdot {\bf V}^T$. Note that both $\bf U$ and $\bf V$ are orthogonal matrices and that $\bf S$ is diagonal with positive values $s_i$ on the main diagonal. 
We consider
\begin{equation}
{\bf a} = {\bf V}^T ( {\bf y}_{t} - {\bf x}_{t} ).  \nonumber
\end{equation}
Because $\bf V$ is an orthogonal matrix, the left side of the equation above is 
a rotation of the right side and
the length $||\bf a||$ is the same as $|| {\bf y}_{t} - {\bf x}_{t}||$.
One can write
\begin{equation}
d^2 ( {\bf y}_{t}, {\bf x}_{t}) = \sum_i a_i^2,  \nonumber
\end{equation}
where the $a_i$ are entries of the vector $\bf a$. Since 
\begin{equation}
{\bf y}_{lin,t+1}- {\bf x}_{lin,t+1} = {\bf U} {\bf S} {\bf a}  \nonumber
\end{equation}
and $\bf U$ is again a rotation matrix, one can write
\begin{equation}
d^2({\bf y}_{lin,t+1}, {\bf x}_{lin,t+1}) = \sum_i s_i^2 a_i^2, \nonumber
\end{equation}
where $s_i$ is the $i$-th component of the diagonal matrix $\bf S$, i.e., the $i$-th singular value. 

In the following we define $s_{\max}=\max_i s_i$ and calculate
\begin{eqnarray} 
&&d^2({\bf y}_{lin,t+1}, {\bf x}_{lin,t+1})  \times \nonumber \\
&& \phi_k (d^2({\bf y}_{lin,t+1}, {\bf x}_{lin,t+1}))   \nonumber \\
& = &  (\sum_i s_i^2 a_i^2 ) \times \phi_k (\sum_i s_i^2 a_i^2 ) \nonumber \\
&\leq & (s_{\max}^2 \sum_i a_i^2 ) \times \phi_k (s_{\max}^2 \sum_i a_i^2 ) \nonumber \\
&\leq & (s_{\max}^2 d^2 ( {\bf y}_{t}, {\bf x}_{t})) \times \phi_k (s_{\max}^2 d^2 ( {\bf y}_{t}, {\bf x}_{t}) )
\label{jaeger2}
\end{eqnarray}

Merging eq. \ref{jaeger1} and eq. \ref{jaeger2} results in the inequality
\begin{equation}
d({\bf y}_{t+1},{\bf x}_{t+1}) \leq s_{\max} d({\bf y}_{t}, {\bf x}_{t}) \times ( \phi_k (s_{\max}^2 d^2 ( {\bf y}_{t}, {\bf x}_{t}))^{0.5}.  \nonumber
 \label{update_inequality}
\end{equation}

First, 
assuming $s_{\max}<1$ and since we know $\phi_k\leq1$, we get an exponential decay 
\begin{equation}
d(y_n,x_n) \leq s_{\max}^t d (x_0,y_0).  \nonumber
\end{equation}

This case is handled by Jaeger's initial proof. With regard to an upper limit of the contraction speed (cf. eq. \ref{metric}), one can find 
\begin{equation}
\tau (\epsilon) = \frac{\log \epsilon - \log(d(y_0,x_0))} {\log s_{\max}}.  \nonumber
\end{equation}

If the largest singular value $s_{\max}>1$, then for some type of connectivities (i.e. normal matrices) the largest absolute eigenvalue is also 
larger than $1$ due to the spectral theorem. In this case, the echo state condition is not always fulfilled, which has been shown also by Jaeger. 

What remains is to check the critical case $s_{\max}=1$. Here again one can discuss two different situations (rather two separate phases of the convergence process) 
separately:

If $d^2 ( {\bf y}_{lin,t}, {\bf x}_{lin,t}) > \gamma$, we can write the update 
inequality of eq. \ref{update_inequality} as:
\begin{equation}
d^2({\bf y}_{t+1},{\bf x}_{t+1}) \leq (1-\eta \gamma^\kappa) d^2({\bf y}_{t},{\bf x}_{t}).  \nonumber
\end{equation} 
Thus, for all $\epsilon^2 \ge \gamma$, the slowest decay process can be covered by 
\begin{equation}
\tau (\epsilon) = \frac{2 \log \epsilon - 2 \log( d(y_0,x_0))} {\log(1-\eta \gamma^\kappa)}.  \nonumber
\end{equation}

If $ \epsilon^2 < \gamma$, then eq. \ref{update_inequality} becomes:
\begin{equation}
d^2({\bf y}_{t+1},{\bf x}_{t+1}) \leq d^2({\bf y}_{t}, {\bf x}_{t}) 
(1- \eta d({\bf y}_{t}, {\bf x}_{t})^{2\kappa} ).  \nonumber
\end{equation}

One can replace 
\begin{equation}
q_t=d^2({\bf y}_{t}, {\bf x}_{t})
\end{equation}
and again consider the sequence 
\begin{equation}
q_{t+1} = q_t (1-\eta q_t^\kappa),   \nonumber
\end{equation}
which is discussed in app. \ref{app_sequence_1}. The result there is that the sequence converges faster than

\begin{equation}
q_*(t)=[ \frac{ \eta} {\kappa} t + q_0^{-\kappa} ] ^ {-1/\kappa}.  \label{slowest_convergence}
\end{equation}
Note that, although the Lyapunov exponent (cf. eq. \ref{lyapunov_exp_eq}) of $q_*(t)$ is zero,
the sequence $q_*(t)$ converges in a power law fashion.
Thus,
\begin{equation}
\lim_{t\rightarrow\infty} q(t) = d^2 ({\bf y}_{t}, {\bf x}_{t}) = 0
\end{equation}
and thus ESP has been proven, QED.

Moreover, one can calculate the upper time limit $\tau(\epsilon)$:
\begin{equation}
\tau(\epsilon) = \frac{\kappa} { \eta} (\epsilon^{-2\kappa} -a_0^{-\kappa}).  \nonumber
\end{equation}

\begin{figure}[t]
\begin{center}
\includegraphics[  width=0.78\paperwidth]{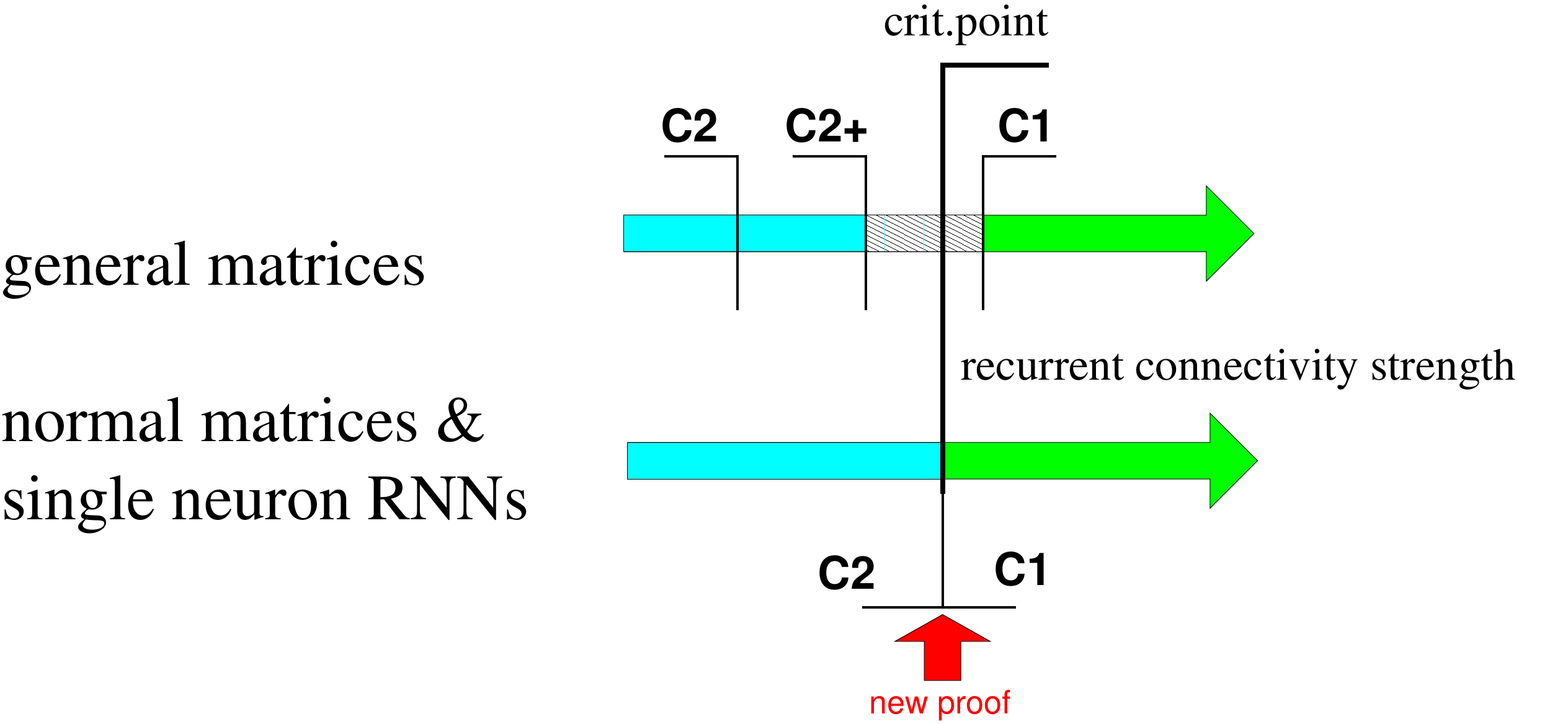} 
\end{center}
\caption{\label{conn_func} The graph illustrates the relation between the different conditions with regard to ESP. {\sf C1} and {\sf C2} are the necessary and sufficient condition according to J\"ager
\cite{jaeger1}. The large arrows represent the connectivity strength of the recurrent synaptic weight matrix $\WWHH$. For a small connectivity strength the ESP is fullfilled (cyan areas). For a strong connectivity, the ESP is not 
fulfilled anymore (green).
 {\sf C2+} represents symbolically the improvement of the sufficient condition according to \cite{tighter,tighter2}. For general matrices there can be a non-zero gap between {\sf C2+} and {\sf C1} which is
drawn diagonally shaded. The transition from ESP to non-ESP happens somewhere within this gap. All three conditions {\sf C1}, {\sf C2} and {\sf C2+} describe an open set. In the case of normal matrices and for one neuron networks the gap is closed except for the separation line itself. The proof of sect. \ref{sec:limit} shows that there it depends on the transfer function if the network is an ESN or not. 
}
\end{figure}

\section{Summary, Discussion and Outlook}
The background of this paper is to investigate the limit of recurrent connectivity in ESNs. The preliminary hypothesis towards the main work can be summarized in fig. \ref{conn_func}. Initially it is hard to quantify the transition point between uniformly state contracting and non-state contracting ESNs exactly. However, for normal matrices and one neuron networks the gap between the sufficient condition and the necessary condition collapses in a way that there are two neighboring open sets. The first open set is known to have the ESP, and the other open set evidently does not have the ESP. What remains is the boundary set. The boundary set is interesting to analyze because it can easily be shown that here power law forgetting can occur.  

The proof of sect. \ref{sec:limit} shows a network is an ESN even if the largest eigenvalue of the recurrent connectivity matrix is equal to $1$ and if the transfer function is either 
eq. \ref{tanh} or \ref{sigmoid}. The proof is also extensible to other transfer functions. On the other hand it is obvious that some transfer functions result in networks that are not ESNs. For example 
a linear transfer function ($\theta(x) = x$) is {\em not} state contracting. 

Even if the network is state contracting, it is not necessarily exponentially uniformly state contracting. Its rate of convergence might follow a power law in the slowest case. Several examples for power law forgetting have been shown in the present work. More examples of preliminary learning have been outlined in \cite{neco2015}. One important target of the present research is to allow for a kind of memory compression in the reservoir by letting only the unpredicted input enter the reservoir.

One ultimate target of the present work is to find a way to organize reservoirs as recurrent filters with a memory compression feature. 
In order to bring concepts of data compression into the field of reservoir computing and in order to project as much as possible of the input history to the limit size reservoir, principles of memory compression have to be transferred into reservoir computing. However, the reservoir computing techniques that are analogous  to classic memory compression have not been identified so far. 

Another topic that needs further investigation is entropy in time series. Power law forgetting is only possible if the time series that relates to the criticality is either of a finite entropy, i.e., from a certain point in time all following entries of the time series can be predicted from the previous entries, or if the network simply ignores certain aspects of the incoming time series.

There also potential analogies in biology. Several measurements of memory decay in humans exist that reveal that there the forgetting follows a power law at least for a large fraction of the investigated examples \cite{rubin1996one,kahana2002note}. 



\section*{Compliance with Ethical Standards}
The author declares that he  has no conflict of interest.

\section*{Ethical approval} 
This article does not contain any studies with human participants or animals performed by the author.

\section*{Acknowledgements}
The Nation Science Council and the Ministry of Science and Technology of Taiwan provided 
the budget for our laboratory. Also thanks go to AIM-HI for various ways of support.

\bibliographystyle{unsrt} 
\bibliography{proof}

\newpage
\appendix
{\Large Appendix}\\

\section{A: Analyze $q_{t+1}=q_t (1- \eta q_t^\kappa)$ \label{app_sequence_1}}
We can consider the sequence $q_t$:
\begin{equation}
q_{t+1}=q_t (1- \eta q_t^\kappa). \label{seq1}
\end{equation}
Convergence can be analyzed in the following way:
\begin{equation}
\Delta q_{t} = q_{t+1} - q_{t} = - \eta q_t^{\kappa+1}. \label{discrete1}
\end{equation} 
Thus, the series $q_{t+1}$ can be written as
\begin{equation}
q_{t+1}= \sum_{t'<t} \Delta q_{t'}  \nonumber
\end{equation}
Thus one comes up with the following discrete formula 
\begin{equation}
\Delta q_{t'} = - \eta q_{t'}^{\kappa+1}. \label{discrete}
\end{equation}
Since $\kappa\geq1$ and $1>\eta>0$, convergence towards null is obvious here. In addition
since the right side of eq. \ref{discrete} is decreasing continuously it is obvious
that it is converging fast than the corresponding solution of the differential equation for 
a function $q_*$:
\begin{equation}
\frac{{\rm d} q_*}{{\rm d} t} = - \eta  q_*^{\kappa+1},  \nonumber
\end{equation}
which is easily solvable to:
\begin{equation}
-\frac{1}{\kappa \eta} q_*^{-\kappa} = t+C,  \nonumber
\end{equation}
where $C$ is the integration constant.
From this solution, by setting $q_0=q_*(0)$ one can derive
\begin{equation}
q_*(t)=[ \frac{ \eta} {\kappa} t + q_0^{-\kappa} ] ^ {-1/\kappa}.  \nonumber
\end{equation}
Note that
\begin{equation}
\lim_{t \rightarrow \infty} q_* = 0.  \nonumber
\end{equation}
Simple algebraic considerations show that $q_*(t)$ covers the sequence $q_t$. 
For a strict proof see app. \ref{sequence_app}.

Thus,
\begin{equation}
q_{*,t} \geq q_t  \nonumber 
\end{equation}
for all $t>0$ if 
\begin{equation}
q_{*,0} = q_0  \nonumber 
\end{equation}

So, the sequences $q_{*,t}$ and $q_t$  converge to zero.

\newpage

\section{\label{sequence_app} B: Sequence $q_{*,t}$ covers up sequence $q_t$ } 

In the following, identical definitions to sect. \ref{sec:limit} are used.
One can start from the statement
\begin{equation}
-\frac{\eta}{\kappa} \geq -  \eta. \label{startappc}
\end{equation}
Since $\eta \geq 0$ and $\kappa \geq 1$, it can easily be seen that the inequality is fulfilled for any combination of 
$s$, $\eta$ and $\kappa$.
One can now extend the numerator and denominator of the right side by $(\frac{\eta}{\kappa} t + C)$ and add also 
$(\frac{\eta}{\kappa} t + C)$ to both sides of the inequality. 
Here and in the following $C$ is defined as
\begin{equation}
C=q_0^{-\kappa} \geq 1.
\end{equation} 
One obtains
\begin{eqnarray}
\frac{\eta}{\kappa} t + C -\frac{\eta}{\kappa} &\geq& \nonumber \\
(\frac{\eta}{\kappa} t + C) - (\frac{\eta}{\kappa} t + C) 
\frac{\eta}{\frac{\eta}{\kappa} t + C}. &&
\end{eqnarray}
A rearrangement of the right side results in
\begin{eqnarray}
\frac{\eta}{\kappa} t + C -\frac{\eta}{\kappa} &\geq& \nonumber \\
(\frac{\eta}{\kappa} t + C) \times (1 - \frac{\eta}{\frac{\eta}{\kappa} t + C}). &&
\end{eqnarray}
One can add $\frac{\eta}{\kappa}$ to both sides and obtain
\begin{eqnarray}
\frac{\eta}{\kappa} t + C  &\geq& \nonumber \\
\frac{s \eta}{\kappa}+ 
(\frac{\eta}{\kappa} t + C) \times (1 -
\frac{\eta}{\frac{s \eta}{\kappa} t + C}). \label{startappc2}
\end{eqnarray}
Now, one can use the fact that
\begin{equation}
1 \geq 1- \frac{\eta}{\frac{\eta}{\kappa} t + C} \geq 
\left(1- \frac{\eta}{\frac{\eta}{\kappa} t + C} \right)^\kappa  \nonumber
\end{equation}
and 
\begin{equation}
\frac{\eta}{\kappa} \geq \frac{\eta}{\kappa} \times \left(1- \frac{\eta}{\frac{\eta}{\kappa} t + C} \right) \geq 
\frac{\eta}{\kappa} \times \left(1- \frac{\eta}{\frac{\eta}{\kappa} t + C} \right)^\kappa  \nonumber
\end{equation}

and thus rewrite inequality eq. \ref{startappc2} as

\begin{eqnarray}
\frac{\eta}{\kappa} t + C &\geq& 
(\frac{\eta}{\kappa} t + C + \frac{\eta}{\kappa}) \times 
\left(1- \frac{\eta}{\frac{\eta}{\kappa} t + C} \right)  \nonumber
\end{eqnarray}

and finally arrive at

\begin{eqnarray}
\left(\frac{\eta}{\kappa} t + C \right) &\geq& 
\left(\frac{\eta}{\kappa} t + C + \frac{\eta}{\kappa} \right) \times 
\left(1- \frac{\eta}{\frac{\eta}{\kappa} t + C} \right)^\kappa.  \nonumber
\end{eqnarray}
One can multiply both sides by
$\left(\frac{\eta}{\kappa} t + C +\frac{\eta}{\kappa} \right)^{-1} \times \left(\frac{\eta}{\kappa} t + C \right)^{-1} $.
So,

\begin{eqnarray}
\left(\frac{\eta}{\kappa} t + C + \frac{\eta}{\kappa} \right)^{-1} &\geq& 
\left(\frac{\eta}{\kappa} t + C \right)^{-1} \times 
\left(1- \frac{\eta}{\frac{\eta}{\kappa} t + C} \right)^\kappa.  \nonumber
\end{eqnarray}
Taking the $k$th root on both sides one gets
\begin{eqnarray}
\left(\frac{\eta}{\kappa} t + C + \frac{\eta}{\kappa} \right)^{-1/\kappa} &\geq& 
\left(\frac{\eta}{\kappa} t + C  \right)^{-1/\kappa} \times   
\left(1- \frac{\eta}{\frac{\eta}{\kappa} t + C} \right).  \nonumber
\end{eqnarray}

One can rearrange the inequality to:

\begin{eqnarray}
\left(\frac{\eta}{\kappa} t + C + \frac{\eta}{\kappa} \right)^{-1/\kappa} &\geq& 
\left(\frac{\eta}{\kappa} t + C  \right)^{-1/\kappa} \times 
\left(1-  \eta \left(\frac{\eta}{\kappa} t + C \right)^{-1} \right) \nonumber \\
\left(\frac{\eta}{\kappa} (t + 1)+ C \right)^{-1/\kappa} &\geq& 
\left(\frac{\eta}{\kappa} t + C  \right)^{-1/\kappa} \times \nonumber \\
& & \left(1- \eta 
\left( \left(\frac{ \eta}{\kappa} t + C \right)^{-1/\kappa} \right)^\kappa \right) 
\end{eqnarray}
Using the definitions of $q_{*,t}$ and $q_{*,t+1}$, i.e.
\begin{equation}
q^*(t)=[\frac{\eta} {\kappa} t + q_0^{-\kappa} ] ^ {-1/\kappa},  \nonumber
\end{equation}

one has finally

\begin{eqnarray}
q_{*,t+1} &\geq& q_{*,t} \times \left(1-\eta (q_{*,t})^\kappa \right).
\end{eqnarray}
proven as a true statement.
Thus,
\begin{equation}
q_{*,t} \geq q_t \geq  d ( {\bf y}_{t}, {\bf x}_{t} )  \nonumber
\end{equation}
for all $t>0$ if 
\begin{equation}
q_{*,0} = q_0 =  d ( {\bf y}_{0}, {\bf x}_{0}).  \nonumber
\end{equation}
Thus, the positive definite sequence $q_{*,t}$ covers $d ( {\bf y}_{t}, {\bf x}_{t})$.

\section{C: Weak contraction with the present transfer function \label{app_contraction_func}}

In this appendix a test function of the form of eq. \ref{weakcontrf}, is verified for the function of eq. \ref{sigmoid} and hyperbolic tangent.
Within this section we test the following values
\begin{eqnarray}
\eta  \,=\,  \frac{1}{48\times n^2},
\gamma \, = \, \frac{1}{2} \; \mbox{and} \;
\kappa \, = \, 2, \label{paramscover}
\end{eqnarray}
for both transfer functions and square norm ($||.||_2$), and $n$-neurons. In order to derive these values, one can start by considering
linear responses $||{\bf y}_{lin,t}-{\bf x}_{lin,t}||$ and the final value $||{\bf y}_{t}-{\bf x}_{t}||$  within one single neuron as
\begin{eqnarray}
{\bf y}_{t} &=& \theta({\bf y}_{lin,t}) \nonumber \\
\mbox{and} \; \; {\bf x}_{t} &=& \theta({\bf x}_{lin,t}).
\end{eqnarray}
We can define
\begin{eqnarray}
y_{lin,t,i} &=& \Delta_i + \zeta_i, \nonumber \\
x_{lin,t,i} &=& \zeta_i, \nonumber \\
\mbox{and} \; \; x_{t,i} - y_{t,i} &=& \delta_i 
\label{linlin}
\end{eqnarray}
where $i$ is the index of the particular hidden layer neuron.
Note that
\begin{equation} 
|| {\bf y}_{lin,t}-{\bf x}_{lin,t} ||^2 = \sum_i \Delta_i^2, \; \; || {\bf y}_{t}-{\bf x}_{t} ||^2 = \sum_i \delta_i^2. \nonumber
\end{equation}

\subsection{One neuron transfer}
In this section considerations are restricted to the case in which one has only one neuron in the hidden layer. For the sake of simplicity, the subscript index $i$ is left out 
in the following considerations. 

Setting in the definitions of eq. \ref{linlin} into the square of eq. \ref{defweak} and for a single neuron and for any $\zeta$ we get
\begin{equation}
 \Delta^2 \; \phi_1(\Delta^2) \geq \left(\theta(\Delta+\zeta) - \theta(\zeta)\right)^2. \nonumber
\end{equation}

Thus, it suffices to consider 
\begin{equation}
\phi_1(\Delta^2) \geq \max_\zeta \omega(\Delta, \zeta),  \label{maxeq}
\end{equation}
where
\begin{equation}
\omega(\Delta, \zeta) = \left(\frac{\theta(\Delta+\zeta) - \theta(\zeta)}{\Delta} \right)^2  \nonumber
\end{equation}
The $\max_\zeta$ can be found by basic analysis. 
Extremal points can be found as solutions of 
\begin{eqnarray}
\frac{\partial}{\partial \zeta} \omega &=& \nonumber \\
\frac{\partial}{\partial \zeta} \left[ \left( \; \frac{\theta(\Delta+\zeta) - \theta(\zeta)}{\Delta}  \right)^2 \right] &=& \nonumber \\
 \frac{2 (\theta(\Delta+\zeta) - \theta(\zeta) ) \times (\dot{\theta}(\Delta+\zeta) - \dot{\theta}(\zeta))}{\Delta^2} &= 0.& 
\end{eqnarray}
This can only be fulfilled if
\begin{equation}
\dot{\theta}(\Delta+\zeta) - \dot{\theta}(\zeta) = 0  \nonumber
\end{equation}
Since $\theta'$ is an even function for both suggested transfer functions, one gets $\zeta=-z/2$ as the extremal point. 
Fundamental analysis shows that this point in both cases is a maximum. Thus, requiring 
\begin{equation}
\phi_1(\Delta^2) \geq \left[\frac{2\theta(\Delta/2)}{\Delta} \right]^2 \label{maxphi}
\end{equation}
also would satisfy eq. \ref{maxeq}.

Numerically, one can find parameters for $\phi_1$ that are $\eta=1/48$ and $\kappa=2$ of as in eq. \ref{paramscover}.

For $\Delta^2>\gamma$, it suffices to check if 
\begin{equation}
\frac{4 {\tanh}^2(\Delta/2)}{\Delta^2} \leq 1 - \eta \gamma ^\kappa  \nonumber
\end{equation}
First the inequality is fulfilled for $\Delta^2=\gamma$. Since the left side of the equation above is strictly decreasingfulfilled for all
values $\Delta^2>\gamma$.

Analogous considerations lead to the same parameters to cover up the transfer function from 
eq. \ref{sigmoid}.

\subsection{Multi-neuron parameters}

For several neurons one has to consider the variational problem of all possible combinations of values of $\Delta_i$. One can start from the proven relation
from the previous section,
\begin{equation}
\delta_i^2 \leq \Delta_i^2 \phi_1 \left( \Delta_i^2 \right)  \nonumber
\end{equation}
Thus,
\begin{equation}
\sum_i \delta_i^2 \leq \sum_i \left( \Delta_i^2 \phi_1 \left( \Delta_i^2 \right) \right),  \nonumber
\end{equation}
implying that
\begin{equation}
\sum_i \left( \Delta_i^2 \phi_1 \left( \Delta_i^2 \right) \right) \leq \sum_i \Delta^2_i - \Delta^2_{max} + \Delta^2_{max} \phi_1 (\Delta^2_{max}),
\label{maxmax}
\end{equation}
where $\Delta^2_{max}= \max_i \Delta^2_i$. The smallest possible $\Delta^2_{max}$ is
\begin{equation}
\frac{\sum_i \Delta^2_i}{n},  \nonumber
\end{equation}
Substituting  into eq. \ref{maxmax}, we get
\begin{eqnarray}
\sum_i \Delta^2_i - \Delta^2_{max} + \Delta^2_{max} \phi_1 (\Delta^2_{max}) &\leq& \nonumber \\
\sum_i \Delta^2_i - \frac{\sum_i \Delta^2_i}{n} + \frac{\sum_i \Delta^2_i}{n} \phi_1 (\frac{\sum_i \Delta^2_i}{n}) &=& \nonumber \\ 
(\sum_i \Delta^2_i) \phi_1 (\frac{\sum_i \Delta^2_i}{n^2}).  \nonumber
\end{eqnarray}
Thus, the inequality (which is equivalent to eq. \ref{defweak}),
\begin{equation}
\sum_i \delta_i^2 \leq \left( \sum_i \Delta_i^2 \right) \phi_k \left( \sum_i \Delta_i^2 \right),  \nonumber
\end{equation}
is fulfilled if $\phi_k$ is defined as
\begin{equation}
\phi_k(x) = \phi_1 \left( \frac{x}{n^2} \right).  \nonumber
\end{equation}
Thus, the parameters from eq. \ref{paramscover} fulfill the inequality of eq. \ref{defweak}.

\end{document}